# JUSTIFYING THE PRINCIPLE OF INTERVAL CONSTRAINTS

Richard E. Neapolitan
Department of Information Science
Northeastern Illinois University
Chicago, Illinois 60625

James Kenevan
Department of Computer Science
Illinois Institute of Technology
Chicago, Illinois 60616

When knowledge is obtained from a database, it is only possible to deduce confidence intervals for probability values. With confidence intervals replacing point values, the results in the set covering model include interval constraints for the probabilities of mutually exclusive and exhaustive explanations. The Principle of Interval Constraints ranks these explanations by determining the expected values of the probabilities based on distributions determined from the interval constraints. This principle was developed using the Classical Approach to probability. This paper justifies the Principle of Interval Constraints with a more rigorous statement of the Classical Approach and by defending the concept of probabilities of probabilities.

## 1. INTRODUCTION

When knowledge is obtained from a database, one can use simple proportions to represent probability values. For example, in the STROKE expert system [1] the set covering model [2] was used to determine mutually exclusive and exhaustive treatment groups, called explanations, from the diseases. In order to rank the explanations the odds method was applied to point probability values obtained from a database [3]. However, although it appeared that the independence assumptions in the odds method, namely that diseases are independent in the subspace that a treatment occurs and in the subspace that it does not, are valid for this application, the results were poor. When a Weighted Kappa Statistic [4] was used to compare the treatments determined by the expert system with those determined by a physician, the correlation was the same as could be expected by chance. A possible reason for these results is that the use of proportions to represent point values misrepresents the 'knowledge' in the data base.

For example, suppose the database contained 500 cases. Of those 500 cases, 10 had both diseases m1 and m2. Of those cases, 1 was administered treatment e1 while 9 were administered the alternative treatment e2. No other cases had either disease or were administered either treatment. If simple proportions are used to determine the point probabilities, $P(ei)$ and $P(ei|mi)$, and the odds method is used to compute the conditional probabilities of e1 and e2 given that both m1 and m2 occurred, the following values are obtained:

$$P(e1|m1,m2) = .86 \quad P(e2|m1,m2) = .99.$$

However, for the subpopulation in the database, the values of these conditional probabilities are .1 and .9 respectively. These values are so different because the independence assumptions clearly are not valid in the subpopulation in the database. However, since the number of cases administered the first treatment were so small, the proportions involving this treatment may not be accurate approximations of the probabilities for the population at large. Therefore the independence assumptions may still be valid in the population at large. In this case one should accurately represent the knowledge in the database by using statistical inference to obtain confidence intervals for the probabilities. Spiegelhalter [5] has noted that it may also be reasonable to obtain confidence intervals from an expert to represent his doubt in point values.

A method for manipulating confidence intervals using the odds method and the set

266

covering model is described in [6]. It is straightforward to derive a similar technique for manipulating intervals in Reggia and Peng's probabilistic causal set covering model [7]. Quinlan [8] has developed a technique which propagates intervals in an inference network. However, Speigelhalter [5] has noted that it is only straightforward to propagate intervals in causal networks under unrealistic independence assumptions.

When the technique from [6] is applied in the above example the following intervals are obtained:

$$P(e1|m1,m2) \epsilon (.0005,.9999) \quad P(e2|m1,m2) \epsilon (.9780,.9999).$$

A data base contains more information about the original probabilities than just confidence intervals. For example, it contains the expected values. However, when the confidence intervals are manipulated to obtain interval constraints for the probabilities of explanations, the only resultant information about the probabilities of these explanations is the interval constraints. Therefore the explanations can be ranked based only on the information in these constraints. The Principle of Interval Constraints [6] accomplishes this by yielding the expected values of the probabilities of mutually exclusive and exhaustive sentences based on distributions determined from interval constraints on the probabilities. This principle obtains these distributions by applying Laplace's [9] Principle of Indifference to the probability values themselves and using the Classical Approach to probability.

The current paper justifies the development of the Principle of Interval Constraints. First off, the Classical Approach has been crticized due to its inapplicability in many situations and to paradoxes [10,11]. In section 3 of this paper, it is shown that, if the Classical Approach is more rigorously stated, these criticisms are eliminated. Others [12] have criticized the notion of probabilities of probabilities. These criticisms are addressed in section 4. Before this, in section 2, the development of the Principle of Interval Constraints is briefly summarized.

## 2. PRINCIPLE OF INTERVAL CONSTRAINTS

Suppose that we have determined N mutually exclusive and exhaustive sentences, $E_i$, and N confidence intervals, $[a_i,b_i]$, where $0 \leq a_i \leq 1$ and $0 \leq b_i \leq 1$, such that, if $p_i$ is the probability of $E_i$, then $p_i \epsilon [a_i,b_i]$. Since the probabilities must sum to one, it is necessary that

$$\sum_{i=1}^{N} a_i \leq 1 \quad \text{and} \quad 1 \leq \sum_{i=1}^{N} b_i. \qquad (2.1)$$

It is assumed in what follows that these inequalities hold. The following lemma, which is proved in [6], yields tighter intervals:

LEMMA 2.1. If $\sum_{i=1}^{N} p_i = 1$ and $p_i \epsilon [a_i,b_i]$ then for $j = 1,2,...,N$,

$$p_j \geq 1 - \sum_{i \neq j} b_i \quad \text{and} \quad p_j \leq 1 - \sum_{i \neq j} a_i.$$

Therefore it is possible to obtain tighter intervals by setting, for $j = 1,2,...,N$,

$$\text{new } b_j \equiv \min(b_j, 1 - \sum_{i \neq j} a_i) \quad \text{and} \quad \text{new } a_j \equiv \max(a_j, 1 - \sum_{i \neq j} b_i).$$

267

In what follows it is assumed that the tighter intervals have been computed so that the following two inequalities hold for j=1,2,...,N:

$$b_j \leq 1 - \sum_{i \neq j} a_i \quad \text{and} \quad a_j \geq 1 - \sum_{i \neq j} b_i. \quad (2.2)$$

The problem now is to determine the expected values of $p_i$ from the intervals $[a_i, b_i]$ given that $p_i \in [a_i, b_i]$ and the points $a_i$ and $b_i$ satisfy (2.1) and (2.2).

Consider first the simple case of three possible events. For a given value $x_1 \in [a_1, b_1]$, the intervals $[a_2, b_2]$ and $[a_3, b_3]$ determine the number of ways that $x_1$ can be the value of $p_1$, since if $x_1$ is a possible value of $p_1$, there must be some $x_2 \in [a_2, b_2]$ and some $x_3 \in [a_3, b_3]$ such that $x_1 + x_2 + x_3 = 1$. If there exist more combinations of points $x_2 \in [a_2, b_2]$ and points $x_3 \in [a_3, b_3]$ such that $x_1' + x_2 + x_3 = 1$ than there exist such that $x_1'' + x_2 + x_3 = 1$, then, by applying the Principle of Indifference to the probability values, we conclude that $x_1'$ is a more probable value of $p_1$ than is $x_1''$. In this way, the intervals $[a_2, b_2]$ and $[a_3, b_3]$ impose a probability density function, $\mu_1(x_1)$, on the interval $[a_1, b_1]$. The expected value of $p_1$ can be computed from this distribution.

The above argument extends easily to the case where there are N sentences. The theorem for that case follows. First we shall determine the density functions in the case of two sentences to illustrate the principles. If there are two possible sentences, inequalities (2.2) imply that

$$a_1 + b_2 = 1 \quad \text{and} \quad a_2 + b_1 = 1.$$

This implies that if $x_1 \in [a_1, b_1]$ then $1 - x_1 \in [a_2, b_2]$, and $p_1$ can have value $x_1$ in exactly one way. Therefore all values in $[a_1, b_1]$ are equally probable to be the value of $p_1$, and the distribution is uniform.

Next the concepts outlined above are extended to N sentences.

**THEOREM 2.1. PRINCIPLE OF INTERVAL CONSTRAINTS.** Suppose there are N mutually exclusive and exhaustive sentences $E_i$, each with $p_i = P(E_i)$, and all that is known is that $p_i \in [a_i, b_i]$ for each i. Then there exist N density functions, $\mu_i(x_i)$, each defined on $[a_i, b_i]$ and each determining the expected value of $p_i$ based on the current information. The intervals $[a_j, b_j]$ for $j \neq i$ determine $\mu_i(x_i)$ by constraining the measure of the ways that, for each $x_i \in [a_i, b_i]$, there exist $x_j \in [a_j, b_j]$ such that $x_i + \sum_{j \neq i} x_j = 1$.

When $N \geq 2$, the function $\mu_i(x_i)$ and the expected value of $p_i$ are obtained from the other intervals as follows (i is fixed at 1 to simplify notation):

1) If $N \geq 3$ define

$$\underline{x}_2(x_1) \equiv \max(a_2, 1-x_1-b_3-b_4-\cdots-b_N)$$
$$\overline{x}_2(x_1) \equiv \min(b_2, 1-x_1-a_3-a_4-\cdots-a_N)$$
$$\underline{x}_3(x_1,x_2) \equiv \max(a_3, 1-x_1-x_2-b_4-\cdots-b_N)$$
$$\overline{x}_3(x_1,x_2) \equiv \min(b_3, 1-x_1-x_2-a_4-\cdots-a_N)$$
$$\vdots$$
$$\underline{x}_{N-1}(x_1,x_2,\ldots,x_{N-2}) \equiv \max(a_{N-1}, 1-x_1-x_2-\cdots-x_{N-2}-b_N)$$
$$\overline{x}_{N-1}(x_1,x_2,\ldots,x_{N-2}) \equiv \min(b_{N-1}, 1-x_1-x_2-\cdots-x_{N-2}-a_N)$$

2) $\hat{p}_1$, the expected value of $p_1$ based on the current information, is given by

$$\hat{p}_1 = \frac{\int_{a_1}^{b_1} \int_{\underline{x}_2}^{\overline{x}_2} \int_{\underline{x}_3}^{\overline{x}_3} \cdots \int_{\underline{x}_{N-2}}^{\overline{x}_{N-2}} \int_{\underline{x}_{N-1}}^{\overline{x}_{N-1}} x_1 \, dx_{N-1} dx_{N-2} dx_{N-3} \cdots dx_2 dx_1}{\int_{a_1}^{b_1} \int_{\underline{x}_2}^{\overline{x}_2} \int_{\underline{x}_3}^{\overline{x}_3} \cdots \int_{\underline{x}_{N-2}}^{\overline{x}_{N-2}} \int_{\underline{x}_{N-1}}^{\overline{x}_{N-1}} dx_{N-1} dx_{N-2} dx_{N-3} \cdots dx_2 dx_1}$$

Notice that the expected probability is a coordinate for the centroid of a convex region inside an N−1 dimensional unit cube, but that it is a different region for each expected probability.

This theorem is proved in [6]. Some examples follow:

**Example 1.** $[a_i, b_i] = [0, b]$ for $i = 1, \ldots, N$ where $1/N \leq b \leq 1$.

Each $\hat{p}_i$ is determined to be $1/N$.

**Example 2.** $[a_1, b_1] = [.2, .4]$, $[a_2, b_2] = [.3, .4]$, $[a_3, b_3] = [.3, .5]$.
The following values are obtained:
$$\hat{p}_1 = .2778000, \quad \hat{p}_2 = .3444667, \quad \hat{p}_3 = .3777333.$$
The expected probabilities sum to 1 as they should.

**Example 3.** $[a_1, b_1] = [.0005, .9999]$, $[a_2, b_2] = [.9780, .9995]$.
The following values are obtained:
$$\hat{p}_1 = .0113, \quad \hat{p}_2 = .9887.$$
These are the intervals from the example given in the introduction. Now the common treatment is ranked much higher than the rare one.

The Principle of Interval Constraints was developed by using the Classical Approach to probability. In the next section criticisms of this approach are eliminated when the approach is stated more rigorously.

## 3. THE CLASSICAL APPROACH

One of the early developments of the foundations of probability theory was the Classical Approach of Laplace [9]. In regards to games of chance, he developed the Principle of Indifference which states that, if an experiment is to be performed, and A and B are two sentences whose truth values are uniquely determined by the outcome of the experiment, and if one's information about the experiment gives one no reason to distinquish the occurrence of A from that of B, then A and B are said to be 'equipossible' and the probability of A is equal to the probability of B. In the Classical Approach Laplace used this principle to construct probability spaces as follows:

*'The theory of chance consists in reducing all the events of some kind to a certain number of cases equally possible, that is to say, to such as we may be equally undecided about in regard to their existence, and in determining the number of cases favorable to the event whose probability is sought. The ratio of this number to that of all the cases possible is the measure of this probability.'*[9]

It is a simple matter to derive the axioms of probability theory based on this construction of a probability space. Furthermore, this is the way we commonly assign simple probabilities. Regardless, the Principle of Indifference and the Classical Approach have become perhaps the most criticized concepts in probability theory. This section briefly reviews these criticisms and shows how a reformulated statement of the Classical Approach overcomes some of them. This subject is discussed in more detail in [13].

The first criticism is the argument that the Classical Approach has no meaning in many standard applications of probability. For example, if one is using the value of .001 to mean the probability that a man entering his 39th year will die in the coming year, there are no equipossible alternatives. Von Mises [14] thus defined probability as only having meaning in the case of an experiment which can be repeated, and he defined probability as a limiting frequence. Based on this definition Von Mises was able to deduce the axioms of probability theory. Since the convergence to which Von Mises refers cannot exist in a pure mathematical sense, the axioms were later proved assuming convergence only in the sense of the weak law of large numbers [15].

Many feel that probabilities used to represent personal belief cannot be defined in terms of the Principle of Indifference or as limiting frequencies. By defining the probability of a sentence E ending up true as the fraction of a whole unit value which one would feel is the fair amount to exchange for the promise that one would receive a whole unit value if E turns out to be true and 0 units if E turns out to be false, de Finetti [16] was able to prove that, unless the axioms of probability theory are valid, a person would have to agree to a bet that he is sure to lose. This has been called the Dutch Book Theorem [17].

This criticism that the Classical Approach cannot be used in the above situations is irrelevant to whether its use is valid in cases where it is possible to it. Few would wait to see 1000 cards drawn from a deck before determining the probability of drawing a heart.

The second criticism is that the Classical Approach is based on logic alone [10]. That is, if we expect the probability of an event to approximate the relative frequency with which the event will occur in repeated trials, the Classical Approach gives no proof that probabilities obtained with this approach will agree with those expectations. Our reason just indicates that they should. This criticism is quite valid, and, indeed, if in 10000 draws hearts came up 1/2 of the time, most would no longer claim that the probability of a heart is still 1/4. However, in practice one must base probability on the available information.

In the absence of empirical data, the application of our reason to the information is the most we can do.

The third criticism is that the Principle of Indifference defines probability in a circular fashion. That is, sentences are equiprobable because that are equipossible. Many [18] feel that Keynes [19] overcame both this criticism and the one based on paradoxes (to be discussed next) in his reformulation of the Principle of Indifference. In [13] it is argued that Keynes does not accomplish this. Briefly, Keynes was unable to eliminate circular reasoning because, in the absence of empirical data, the judgement of equiprobable or equipossible alternatives can only be based on our reason. It cannot be proved. In practice our reason determines that 'indivisible' alternatives are equipossible. The concept of indivisible alternatives is discussed at length in [18] and [19].

The last criticism is that in the case of paradoxes the Classical Approach leads to contradictory results. Since it is well-known, one of Bertrand's geometrical paradoxes will be used for illustration. In this paradox, a line is dropped at random on a circle of radius one. The problem is to determine the probability that the length of the chord produced by the line is greater than one. It is a simple matter to show [13] that if the Principle of Indifference is applied to a rectangular coordinate a value of $\sqrt{3}/2$ is obtained for this probability, while if it is applied to the angle in polar coordinates a value of 2/3 is obtained. It appears that we have different probabilities for the same event depending on the coordinate system used.

To explain why such paradoxes are not antinomies we must first review the relationship between probability and information. For example, in the draw of the top card from a deck, if one had only the information that the deck were standard the probability of drawing the ace of hearts would be 1/52, while if one peaked at the bottom card and had the information that it was a king, the probability of drawing that ace would be 1/51. Some argue that some probabilities are metaphysically 'real' while others are due only to lack of information. They claim a coin possesses a 'real' probability of turning up heads based on the coin's composition. Others believe in determinism in the universe. They maintain that if a 'demon' knew the position and momentum of every particle in the universe, he could predict every future event [9]. If one knew the position of the coin in the hand, the torque placed on the coin, etc, one would know for certain whether a heads would come up. This belief conflicts with the modern theory of quantum mechanics. Some physicists [20] are unsatisfied with quantum mechanics for that reason. This issue is discussed at length in [18]. Regardless of ones view on this issue, in practice both the gambler and the expert must base their probabilities on the available information.

Next it is shown that, if the Classical Approach is rigorously stated, the information in the paradoxes is not sufficient to use this approach. Therefore the paradoxes are not antinomies because they are not situations in which the approach should be used. The Principle of Indifference states that if the information gives one no reason to choose one alternative over the other they should be assigned the same probability. The Classical Approach derives probability spaces by applying this principle to a set of mutually exclusive and exhaustive sentences. However, this approach violates the Principle of Indifference by not requiring that any two sentences, for which the information gives one no reason to judge one more probable than the other, be assigned the same probabilities in the resultant probability space. The contradictions in all the paradoxes arise because, in a derived probability space, two such sentences are not assigned the same probability. Let us then reformulate the Classical Approach as follows:

*Let an experiment be given and let F be the set of all sentences whose truth values are uniquely determined by the outcome of the experiment. Suppose it is possible to create a*

*probability measure on F such that for every $A,B \in F$, for which the information in the experiment gives one no reason to judge either A or B more probable than the other, the probability of A is equal to the probability of B. Then the resultant probability space satisfies the criteria of the Principle of Indifference. If not, then the information is not sufficient to create a probability measure on F based on the Principle of Indifference.*

In practice the distribution is obtained from a set of mutually exclusive and exhaustive sentences. It does not matter whether these sentences are judged to be equiprobable or whether it is only not possible to judge that one is more probable than the other. We are not in quest of metaphysically real probabilities. Rather we seek a method for determining probability values from the available information which are most in accord with our reason. If we are able to assign equal probabilities to every two sentences, for which the information does not imply one is more probable than the other, then we have a probability measure which has not assumed any additional information. On the other hand if we create a probability measure which assigns different probabilities to two such sentences then we are assuming additional information, namely that their probabilities are not equal. If these are the only probability measures that can be created, then the information is not sufficient to create a probability measure based on the Principle of Indifference. We will now see that this is the problem in the paradoxes.

Looking again at Bertrand's Paradox, the statement of the experiment is that the line is dropped at 'random'. Since this information says nothing to differentiate one angle's occurrence from another's, it is not possible to judge one angle more probable than another and therefore all angles must be given the same probability. When all angles are given equal probabilities, larger values of x are assigned larger probabilities than smaller values of x. However, since the original information says nothing to differentiate one x value's occurrence from another's, this violates the fact that we'd not judge one value of x to be more probable than the other. A similar problem concerning angles results when rectangular coordinates are used to solve the problem. Since all angles can have the same probability only if all x coordinates do not, it is not possible to create a probability measure based on the Principal of Indifference.

Therefore the Classical Approach cannot be used in this problem. There is nothing contradictory about this. As Keynes [19] said, the Classical Approach cannot be used on every body of information. For example, suppose the information is that an urn contains one white ball and the remaining balls are black, and we wish to compute the probability of drawing a black ball. No one would deem this information sufficient to apply the Classical Approach. However, the reason it is insufficient is because of the condition in the reformulated Classical Approach. Based on the information, there is no reason to judge that a given number of black balls is more probable than any other number. However, it is not possible to create a probability measure in which the sentence 'the bin contains n black balls' has the same probability for every n.

It may seem that the Classical Approach was reformulated just to eliminate the paradoxes, but that was not the case. We have reformulated the Classical Approach to bring the application of the Principle of Indifference in this approach in line with our reason. To see this we shall analyze Bertrand's paradox further. If the information were more precise it would be possible to use the Classical Approach. For example, if the information were that the hand is randomly moving to the left or right, then the solution using rectangular coordinates would be correct. As the problem was stated, we can only conclude that the hand is moving in a totally random fashion. However, the movement of a hand in a random fashion is not a well defined concept. It makes sense that such information is not sufficient to determine a probability measure.

The purpose of this section has not been to return the Classical Approach to its status of the 18th century. Rather it has been to show that, in the absence of empirical information, a cautious application of this approach yields probability values which agree with our reason.

Since the Principle of Interval Constraints was based on an application of the reformulated Classical Approach, the development of this principle contains no contradictions. There remains, however, the question of whether it makes sense to assign probabilities to probability values.

## 4. PROBABILITIES OF PROBABILITIES

Pearl [12] states that 'traditional probability theory insists that probabilities be assigned strictly....to sentences whose truth values can, at least in principle be verified unequivocally by emperical test....The truth of probabilistic statements about any specific event cannot be ascertained nor falsified emperically; once we observe the uncertain event, the probability becomes either one or zero'. Pearl thus concludes that it makes no sense to discuss probabilities of probabilities.

This problem is resolved by distinguishing the experiment which determines the probability value from the experiment which uses the probability value. Consider the following example: Create a large set of urns such that possible proportions of type A and B items are equally represented in the set. The act of selecting an urn is the experiment which determines a probability value while the act of picking an item from that urn is the experiment which uses that probability value. Pearl does acknowledge such two–level lottery processes. He dismisses them by arguing that the betting behavior of a rational agent believing in such a process cannot be distinquished from that of an agent believing in an equivalent single–level model. His argument, however, does not invalidate the example. The example shows that the outcome of an experiment can determine a probability value.

Continuing the example with the urns further, now set the compositions of the urns so that the proportions of type A items and type B items are uniformly distributed in [0,.5] and [.5,1] respectively. Next randomly pick a large number of urns. If n is the number of urns picked and $p_i$ is the actual proportion of type A items in the ith urn picked, then .25 is the expected value of each $p_i$. The strong law of large numbers therefore implies that, with probability 1,

$$\lim_{n\to\infty} \frac{p_1+p_2+\cdots+p_n}{n} = .25 ,$$

and it is 'almost certain' that the proportion of type A items in all the urns collectively can be made arbitrarily close to .25 by taking n sufficiently large. Therefore if a large number of items are drawn from each urn picked, the strong law of large numbers again implies that it is 'almost certain' that the proportion of type A items drawn is close to .25. Thus, if when each item is drawn, we bet in accordance with the expected value obtained using the Principle of Interval Constraints, we should approximately break even in the long run.

Such an example is a bit contrived. Another example illustrates where the probability values are distributed in nature. Suppose two individuals race and we have no information as to their relative speeds. The constraints on the probability of one individual winning are that the probability is a value in the interval [0,1] and that there are exactly two alternatives. Choosing those two individuals is the experiment which uniquely determines that probability. It may not be possible to compute this probability exactly. However, by using probability theory in this situation, one is hypothesizing that there is a probability value associated with certain information about the individuals. In principle then a demon could know it, and one could determine the value exactly by consulting the demon. In practice one can closely approximate the value by observing them race many times and using statistical inference. To obtain another probability value subject to these same constraints one must pick two other individuals to race, toss a coin about which one has no information as to the probability of heads, or pick any other event with two possible outcomes and no other information. The Principle of Interval Constraints hypothesizes that the probabilities in these experiments are distributed in nature according to the Principle of Indifference. If this hypothesis is correct, the example with the urns shows that the most rational betting behavior is to bet according to the expected values determined by the Principle of Interval Constraints.